\documentclass[10pt,twocolumn]{article}

\usepackage{cvpr}
\usepackage{times}
\usepackage{epsfig}
\usepackage{graphicx}
\usepackage{amsmath}
\usepackage{amssymb}
\usepackage{amsfonts}
\usepackage{bm}
\usepackage[table,xcdraw]{xcolor}
\usepackage{mathtools}
\usepackage{booktabs} 
\usepackage[utf8x]{inputenc} 
\usepackage{caption,subcaption}

\usepackage[pagebackref=true,breaklinks=true,colorlinks,bookmarks=false]{hyperref}

\cvprfinalcopy 

\def\cvprPaperID{****} 

\usepackage{soul}
\usepackage{xcolor}
\usepackage{amsmath}

\newcommand{\mat}[1]{\bm{#1}}

\newcommand{\arr}[1]{\mat{\mathcal{#1}}}

\def\figurePath{images/}
\def\myfigure#1#2{\begin{figure}[t]\centering\includegraphics*[width = \linewidth]{\figurePath#1}\caption{#2}\label{fig:#1}\vspace{-5pt}\end{figure}}

\def\mycfigure#1#2{\begin{figure*}[t]\centering\includegraphics*[clip, width = \linewidth]{\figurePath#1}\caption{#2}\label{fig:#1}\vspace{-6pt}\end{figure*}}

\def\myparagraph#1{\vspace{2mm}\noindent\textbf{#1.}}

\newcommand{\refTbl}[1]{Table~\ref{tbl:#1}}

\definecolor{unsurecolor}{rgb}{1,.85,.7}
\definecolor{changedcolor}{rgb}{.85,1,.7}

\soulregister\ref7
\soulregister\cite7
\soulregister\citeetal7
\soulregister\etal0
\soulregister\eg0
\soulregister\scene1
\soulregister\refTbl1

\DeclareGraphicsExtensions{.pdf,.ai,.psd,.jpg,.png}
\DeclareMathOperator*{\argmin}{argmin}

\ifcvprfinal\fi
\begin{document}

\title{LatentFusion: End-to-End Differentiable Reconstruction and Rendering \\for Unseen Object Pose Estimation}

\author{
Keunhong Park\textsuperscript{1,2}\footnotemark\qquad 
Arsalan Mousavian\textsuperscript{2}\qquad 
Yu Xiang\textsuperscript{2}\qquad 
Dieter Fox\textsuperscript{1,2} \\
\textsuperscript{1}University of Washington\qquad 
\textsuperscript{2}NVIDIA
}



\twocolumn[{
\renewcommand\twocolumn[1][]{#1}
\maketitle
\begin{center}
  \vspace{-5mm}
  \centering
  \includegraphics[width=\linewidth]{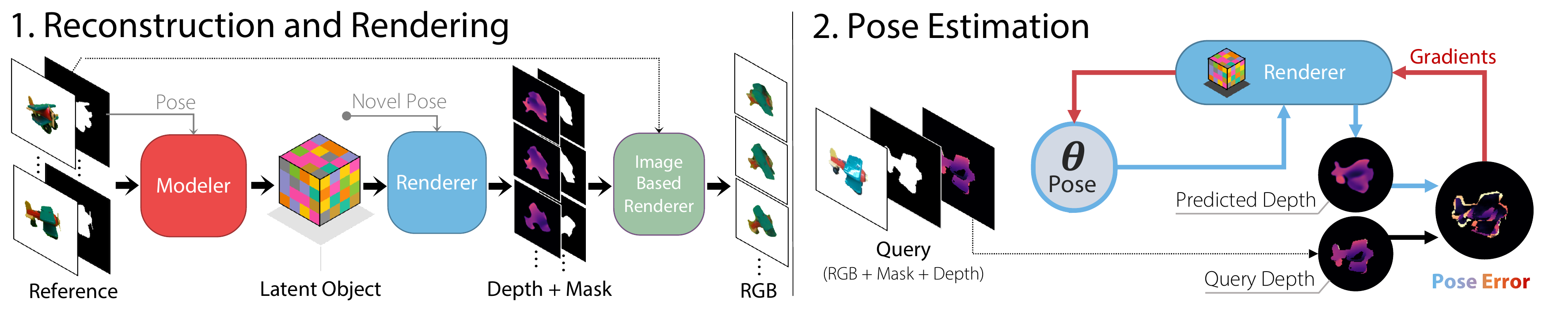}
  \vspace{-4mm}
\captionof{figure}{We present an end-to-end differentiable reconstruction and rendering pipeline. We use this pipeline to perform pose estimation on unseen objects using simple gradient updates in a render-and-compare fashion.}
    \label{:teaser}
\end{center}}]


\begin{abstract}


Current 6D object pose estimation methods usually require a 3D model for each object. These methods also require additional training in order to incorporate new objects. As a result, they are difficult to scale to a large number of objects and cannot be directly applied to unseen objects.

We propose a novel framework for 6D pose estimation of \emph{unseen} objects. We present a network that reconstructs a latent 3D representation of an object using a small number of reference views \emph{at inference time}. Our network is able to render the latent 3D representation from arbitrary views. Using this neural renderer, we directly optimize for pose given an input image. By training our network with a large number of 3D shapes for reconstruction and rendering, our network generalizes well to unseen objects. We present a new dataset for unseen object pose estimation--MOPED. We evaluate the performance of our method for unseen object pose estimation on MOPED as well as the ModelNet and LINEMOD datasets. Our method performs competitively to supervised methods that are trained on those objects. Code and data will be available at https://keunhong.com/publications/latentfusion/.

\end{abstract}
\vspace{-4mm}


\footnotetext[1]{Work done while author was an intern at NVIDIA.}

\section{Introduction}

The \emph{pose} of an object defines where it is in space and how it is oriented.
An object pose is typically defined by a 3D orientation (rotation) and translation comprising six degrees of freedom (6D). Knowing the pose of an object is crucial for any application that involves interacting with real world objects. For example, in order for a robot to manipulate objects it must be able to reason about the pose of the object. In augmented reality, 6D pose estimation enables virtual interaction and re-rendering of real world objects.

In order to estimate the 6D pose of objects, current state-of-the-art methods \cite{xiang2017posecnn,Deng2019,wang2019densefusion} require a 3D model for each object. Methods based on renderings~\cite{tremblay2018deep} usually need high quality 3D models typically obtained using 3D scanning devices. Although modern 3D reconstruction and scanning techniques such as \cite{newcombe2011kinectfusion} can generate 3D models of objects, they typically require significant effort. It is easy to see how building a 3D model for every object is an infeasible task.

Furthermore, existing pose estimation methods require extensive training under different lighting conditions and occlusions. For methods that train a single network for multiple objects \cite{xiang2017posecnn}, the pose estimation accuracy drops significantly with the increase in the number of objects. This is due to large variation of object appearances depending on the pose. To remedy this mode of degradation, some approaches train a separate network for each object \cite{tremblay2018deep,sundermeyer2018implicit,Deng2019}. This approach is not scalable to a large number of objects. Regardless of using a single or multiple networks, all model-based methods require extensive training for unseen test objects that are not in the training set.

In this paper, we investigate the problem of constructing a 3D object representations for 6D object pose estimation without 3D models and without extra training for unseen objects during test time. The core of our method is a novel neural network that takes a set of reference RGB images of a target object with known poses, and internally builds a 3D representation of the object. Using the 3D representation, the network is able to render arbitrary views of the object. To estimate object pose, the network compares the input image with its rendered images in a gradient descent fashion to search for the best pose where the rendered image matches the input image. Applying the network to an unseen object only requires collecting views with registered poses using traditional techniques \cite{newcombe2011kinectfusion} and feeding a small subset of those views with the associated poses to the network, instead of training for the new object which takes time and computational resources.




Our network design is inspired by space carving~\cite{kutulakos2000theory}. We build a 3D voxel representation of an object by computing 2D latent features and projecting them to a canonical 3D voxel space using a \emph{de}projection unit inspired by~\cite{Nguyen-Phuoc2018}. This operation can be interpreted as space carving in latent space. Rendering a novel view is conducted by rotating the latent voxel representation to the new view and projecting it into the 2D image space. Using the projected latent features, a decoder generates a new view image by first predicting the depth map of the object at the query view and then assigning color for each pixel by combining corresponding pixel values at different reference views.

To reconstruct and render unseen objects, we train the network on the ShapeNet dataset~\cite{chang2015shapenet} randomly textured with images from the MS-COCO dataset~\cite{lin2014microsoft} under random lighting conditions. Our experiments show that the network generalizes to novel object categories and instances. For pose estimation, we assume that the object of interest is segmented with a generic object instance segmentation method such as \cite{Xie2019}. The pose of the object is estimated by finding a 6D pose that minimizes the difference between a predicted rendering and the input image. Since our network is a differentiable renderer, we optimize by directly computing the gradients of the loss with respect to the object pose. Fig.~\ref{:teaser} illustrates our reconstruction and pose estimation pipeline.





Some key benefits of our method are:
\begin{enumerate}
    \item \emph{Ease of Capture} -- we perform pose estimation given just a few reference images rather than 3D scans;
    \item \emph{Robustness to Appearance} -- we create a latent representation from images rather than relying on a 3D model with baked appearance; and
    \item \emph{Practicality} -- our zero-shot formulation requires only one neural network model for all objects and requires no training for novel objects.
\end{enumerate}

In addition, we introduce the Model-free Object Pose Estimation Dataset (MOPED) for evaluating pose estimation in a zero-shot setting. Existing pose estimation benchmarks provide 3D models and rendered training images sequences, but typically do not provide casual real-world reference images. MOPED provides registered reference and test images for evaluating pose estimation in a zero-shot setting.

\mycfigure{overview}{A high level overview of our architecture. 1) Our modeling network takes an image and mask and predicts a feature volume for each input view. The predicted feature volumes are then fused into a single canonical \emph{latent object} by the fusion module. 2) Given the latent object, our rendering network produces a depth map and a mask for any camera pose. }

\section{Related Work}
\myparagraph{Pose Estimation} 
Pose estimation methods fall into three major categories. The first category tackles the problem of pose estimation by designing network architectures that facilitate pose estimation \cite{MousavianCVPR17, 3DRCNN_CVPR18, kehl2017ssd}. The second category formulates the pose estimation by predicting a set of 2D image features, such as the projection of 3D box corners \cite{tremblay2018deep, wang2019densefusion, hu2019segmentation, peng2019pvnet} and direction of the center of the object \cite{xiang2017posecnn}, then recovering the pose of the object using the predictions. The third category estimates the pose of objects by aligning the rendering of the 3D model to the image. DeepIM~\cite{Li2018} trains a neural network to align the 3D model of the object to the image. Another approach is to learn a model that can reconstruct the object with different poses \cite{sundermeyer2018implicit, Deng2019}. These methods then use the latent representation of the object to estimate the pose. A limitation of this line of work is that they need to train separate auto-encoders for each object category and there is a lack of knowledge transfer \emph{between} object categories. In addition, these methods require high-fidelity textured 3D models for each object which are not trivial to build in practice since it involves specialized hardware~\cite{singh2014bigbird}. Our method addresses these limitations: our method works with a set of reference views with registered poses instead of a 3D model. Without additional training, our system builds a latent representation from the reference views which can be rendered to color and depth for arbitrary viewpoints. Similar to \cite{sundermeyer2018implicit, Deng2019}, we seek to find a pose that minimizes the difference in latent space between the query object and the test image. Differentiable mesh renderers have been explored for pose estimation~\cite{palazzi2018end, chen2019learning} but still require 3D models leaving the acquisition problem unsolved.

\myparagraph{3D shape learning and novel view synthesis} Inferring shapes of objects at the category level has recently gained a lot of attention. Shape geometry has been  represented as voxels~\cite{choy20163d}, Signed Distance Functions (SDFs)~\cite{park2019deepsdf}, point clouds~\cite{pointflow}, and as implicit functions encoded by a neural network~\cite{Sitzmann2019}. These methods are trained at the category level and can only represent different instances within the categories they were trained on. In addition, these models only capture the \emph{shape} of the object and do not model the \emph{appearance} of the object. To overcome this limitation, recent works~\cite{Nguyen-Phuoc2019, Nguyen-Phuoc2018, Sitzmann2019} decode appearance from neural 3D latent representations that respect projective geometry, generalizing well to novel viewpoints. Novel views are generated by transforming the latent representation in 3D and projecting it to 2D. A decoder then generates a novel view from the projected features. Some methods find a nearest neighbor shape proxy and infer high quality appearances but cannot handle novel categories~\cite{wang2016unsupervised,park2018photoshape}. Differentiable rendering~\cite{Li:2018:DMC, Nguyen-Phuoc2018, liu2019softras} systems seek to implement the rendering process (rasterization and shading) in a differentiable manner so that gradients can be propagated to and from neural networks. Such methods can be used to directly optimize parameters such as pose or appearance. Current differentiable rendering methods are limited by the difficulty of implemented complex appearance models and require a 3D mesh. We seek to combine the best of these methods by creating a differentiable rendering pipeline that does not require a 3D mesh by instead building voxelized latent representations from a small number of reference images. 

\myparagraph{Multi-View Reconstruction} Our method takes inspiration from multi-view reconstruction methods. It is most similar to space carving~\cite{kutulakos2000theory} and can be seen as a latent-space extension of it. Dense fusion methods such as \cite{newcombe2011kinectfusion,whelan2015elasticfusion} generate dense point clouds of the objects from RGB-D sequences. Recent works~\cite{tulsiani2017multi, tulsiani2018multi} have explored ways to learn object representations from unaligned views. These methods recover coarse geometry and pose given an image, but require large amounts of training data for a single object category. Our method builds on both approaches: we train a network to reconstruct an object; however, instead of training per-object or per-category, we provide multiple reference images at inference time to create a 3D latent representation which can be rendered from novel viewpoints.







\section{Overview}
\label{sec:overview}

We present an end-to-end system for novel view reconstruction and pose estimation. We present our system in two parts.  Sec.~\ref{sec:recon} describes our reconstruction pipeline which takes a small collection of reference images as input and produces a flexible representation which can be rendered from novel viewpoints. We leverage multi-view consistency to construct a latent representation and do not rely on category specific shape priors. This key architecture decision enables generalization beyond the distribution of training objects. We show that our reconstruction pipeline can accurately reconstruct unseen object categories from real images. In Sec.~\ref{sec:pose}, we formulate the 6D pose estimation problem using our neural renderer. Since our rendering process is fully differentiable, we directly optimize for the camera parameters without the need for additional training or code-book generation for new objects.

\myparagraph{Camera Model}\label{sec:camera_model} Throughout this paper we use a perspective pinhole camera model with an intrinsic matrix
\begin{align}
    \mat{K} = \begin{pmatrix}
f_u & 0   &  u_0 \\
0   & f_v &  v_0 \\
0   & 0   &  1 
\end{pmatrix},
\end{align}
and a homogeneous extrinsic matrix $\mat{E} = [\mat{R}|\mat{t}]$, where $f_u$ and $f_v$ are the focal lengths, $u_0$ and $v_0$ are the coordinates of the camera principal point, and $\mat{R}$ and $\mat{t}$ are rotation and translation of the camera, respectively. We also define a \emph{viewport} cropping parameter $\mat{c} = (u_-, v_-, u_+, v_+)$ which represents a bounding box around the object in pixel coordinates. For brevity, we refer to the collection of these camera parameters as $\mat{\theta}=\{\mat{R},\mat{t},\mat{c}\}$.

\section{Neural Reconstruction and Rendering}
\label{sec:recon}
Given a set of $N$ reference images with associated object poses and object segmentation masks, we seek to construct a representation of the object which can be rendered with arbitrary camera parameters. Building on the success of recent methods \cite{Nguyen-Phuoc2019, Sitzmann2019}, we represent the object as a latent 3D voxel grid. This representation can be directly manipulated using standard 3D transformations--naturally accommodating our requirement of novel view rendering. The overview of our method is shown in Fig.~\ref{fig:overview}. There are two main components in our reconstruction pipeline:
1) \emph{Modeling} the object by predicting per-view feature volumes and fusing them into a single canonical latent representation;
    2) \emph{Rendering} the latent representation to depth and color images.

\subsection{Modeling}
\label{sec:modeling}

Our modeling step is inspired by space carving \cite{kutulakos2000theory} in that our network takes observations from multiple views and leverages multi-view consistency to build a canonical representation. However, instead of using photometric consistency, we use latent features to represent each view which allows our network to learn features useful for this task.

\myparagraph{Per-View Features} We begin by generating a feature volume for each input view $\mat{\mathcal{I}}_i \in \{\mat{\mathcal{I}}_1,\ldots,\mat{\mathcal{I}}_N\}$. Each feature volume corresponds to the camera frustum of the input camera, bounded by the viewport parameter $\mat{c}=(u_-,v_-,u_+,v_+)$ and depth-wise by $z\in[z_c-r, z_c+r]$ where $z_c$ is the distance to the object center and $r$ is the radius of the object. Fig.~\ref{fig:camera_frustum} illustrates the generation of the per-view features.  Similar to \cite{Sitzmann2018}, we use U-Nets \cite{ronneberger2015u} for their property of preserving spatial structure. We first compute 2D features $g_\text{pix}(\mat{x}_i)\in\mathbb{R}^{C\times H\times W}$ by passing the input $\mat{x}_i$ (an RGB image $\arr{I}_i$, a binary mask $\arr{M}_i$, and optionally depth $\arr{D}_i$) through a 2D U-Net.
The deprojection unit ($p_\uparrow$) then lifts the 2D image features in $\mathbb{R}^{C\times H\times W}$ to 3D volumetric features in $\mathbb{R}^{(C/D)\times D\times H\times W}$ by factoring the 2D channel dimension into the 3D channel dimension $C^\prime=C/D$ and depth dimension $D$. This deprojection operation is the exact opposite of the projection unit presented in \cite{Nguyen-Phuoc2018}. The lifted features are then passed through a 3D U-Net $g_\text{cam}$ to produce the volumetric features for the camera: $\mat{\Phi_i} = g_\text{cam}\circ p_\uparrow\circ g_\text{pix}(\mat{x}_i)\in\mathbb{R}^{C^\prime\times M\times M\times M}$.

\myfigure{camera_frustum}{The $M\times M\times M$ per-view feature volumes computed in the modeling network corresponds a depth bounded camera frustum. The blue box on the image plane is determined by the camera crop parameter $\mat{c}=(u_-,v_-,u_+,v_+)$ and together with the depth determines the bounds of the frustum.} 

\myparagraph{Camera to Object Coordinates}\label{sec:transform} Each voxel in our feature volume represents a point in 3D space. Following recent works \cite{Nguyen-Phuoc2018,Nguyen-Phuoc2019,Sitzmann2018}, we transform our feature volumes directly using rigid transformations.
Consider a continuous function $\phi(\mat{x}) \in\mathbb{R}^{C^\prime}$ defining our camera-space latent representation, where $\mat{x}\in\mathbb{R}^3$ is a point in camera coordinates. The feature volume $\mat{\Phi}$ is a discrete sample of this function. This representation in object space is given by $\psi(\mat{x}^\prime) = \phi(\mat{W}^{-1}\mat{x}^\prime)$ where $\mat{x}^\prime$ is a point in object coordinates and $\mat{W}=[\mat{R}|\mat{t}]$ is an object-to-camera extrinsic matrix. We compute the object-space volume $\mat{\hat\Psi}$ by sampling $\phi(\mat{W}^{-1}\mat{x}^\prime_{ijk})$ for each object-space voxel coordinate $\mat{x}^\prime_{ijk}$. In practice, this is done by trilinear sampling the voxel grid and edge-padding values that fall outside. Given this transformation operation  $T_{\text{c}\rightarrow \text{o}}$, the object-space feature volume is given by
$
\mat{\hat\Psi_i} = 
T_{\text{c}\rightarrow \text{o}}(\mat{\Phi_i}).
$

\myparagraph{View Fusion}\label{sec:fusion}
We now have a collection of feature volumes $\mat{\hat\Psi_i} \in \{\mat{\hat\Psi_i},\ldots,\mat{\hat\Psi_N}\}$, each associated with an input view. Our fusion module $f$ fuses all views into a single canonical feature volume: $ \mat\Psi = f(\mat{\hat\Psi_1},\ldots,\mat{\hat\Psi_N}) $.

Simple channel-wise average pooling yields good results but we found that sequentially integrating each volume using a Recurrent Neural Network (RNN) similarly to \cite{Sitzmann2018} slightly improved reconstruction accuracy (see Sec.~\ref{sec:ablation}). Using a recurrent unit allows the network to keep and ignore features from views in contrast to average pooling. This facilitates comparisons between different views allowing the network to perform operations similar to the photometric consistency criterion used in space carving~\cite{kutulakos2000theory}. We use a Convolutional Gated Recurrent Unit (ConvGRU) \cite{ballas2015delving} so that the network can leverage spatial information. 

\myfigure{fusion_figure}{We illustrate two methods of fusion per-view feature volumes. (1) Simple channel-wise average pooling and (2) a recurrent fusion module similar to that of~\cite{Sitzmann2018}}

\subsection{Rendering}
\label{sec:rendering}

Our rendering module takes the fused object volume $\mat\Psi$ and renders it given arbitrary camera parameters $\mat\theta$. Ideally, the rendering module would directly regress a color image. However, it is challenging to preserve high frequency details through a neural network. U-Nets \cite{ronneberger2015u} introduce skip connections between equivalent-scale layers allowing high frequency spatial structure to propagate to the end of the network, but it is unclear how to add skip connections in the presence of 3D transformations. Existing works such as \cite{Sitzmann2018,Lombardi2019} train a single network for each scene allowing the decoder to memorize high frequency information while the latent representation encodes state information. Trying to predict color without skip connections results in blurry outputs. We side-step this difficulty by first rendering depth and then using an image-based rendering approach to produce a color image.

\myparagraph{Decoding Depth} Depth is a 3D representation, making it easier for the network to exploit the geometric structure we provide. In addition, depth tends to be locally smoother compared to color allowing more information to be compactly represented in a single voxel.

Our rendering network is a simple inversion of the reconstruction network and bears many similarities to RenderNet~\cite{Nguyen-Phuoc2018}. First, we pass the canonical object-space volume $\mat\Psi$ through a small 3D U-Net ($h_\text{obj}$) before transforming it to camera coordinates using the method described in Sec.~\ref{sec:transform}. We perform the transformation with an object-to-camera extrinsic matrix $\mat{E}$ instead of the inverse $\mat{E}^{-1}$. A second 3D U-Net ($h_\text{cam}$) then decodes the resulting volume to produce a feature volume:
$\mat{\Psi^\prime} = h_\text{cam}\circ T_{\text{o}\rightarrow\text{c}}\circ h_\text{obj}(\mat\Psi) $
which is then flattened to a 2D feature grid $\mat{\Phi^\prime} = p_\downarrow(\mat{\Psi^\prime)}$ using the projection unit ($p_\downarrow$) from \cite{Nguyen-Phuoc2018} by first collapsing the depth dimension into the channel dimension and applying a 1x1 convolution. The resulting features are decoded by a 2D U-Net ($h_\text{pix}$) with two output branches for depth ($h_\text{depth}$) and for a segmentation mask ($h_\text{mask}$). The outputs of the rendering network are given by $\mat{y_\text{depth}}(\mat{\Phi^\prime}) = h_\text{depth}\circ h_\text{pix} (\mat{\Phi^\prime})$ and $\mat{y_\text{mask}}(\mat{\Phi^\prime}) = h_\text{mask}\circ h_\text{pix} (\mat{\Phi^\prime})$.

\myparagraph{Image Based Rendering (IBR)} We use image-based rendering~\cite{shum2000review} to leverage the reference images to predict output color. Given the camera intrinsics $\mat{K}$ and depth map for an output view, we can recover the 3D object-space position of each output pixel $(u,v)$ as 
    $\mat{X}=\mat{E}^{-1}\left(\frac{u - u_0}{f_u}z, \frac{v - v_0}{f_v}z, z, 1\right)^T$,
which can be transformed to the input image frame as $\mat{x_i^\prime}=\mat{K_i}\mat{W_i}\mat{X}$ for each input camera $\mat{\theta}_i=\{\mat{K_i}, \mat{W_i}\}$. The output pixel can then copy the color of the corresponding input pixel to produce a reprojected color image. 


The resulting reprojected image will contain invalid pixels due occlusions. There are multiple strategies to weighting each pixel including 1) weighting by reprojected depth error, 2) weighting by similarity between input and query cameras, 3) using a neural network. The first choice suffers from artifacts in the presence of depth errors or thin surfaces. The second approach yields reasonable results but produces blurry images for intermediate views. We opt for the third option. Following deep blending \cite{hedman2018deep}, we train a network that predicts blend weights $\arr{W}_i$ for each reprojected input $\arr{I}^\prime_i$: $\mat{\mathcal{I}}_o = \sum_i{ \arr{W}_i \odot \mat{\mathcal{I}}^\prime_i },$
where $\odot$ is an element-wise product.  The blend weights are predicted by a 2D U-Net. The inputs to this network are 1) the depth predicted by our reconstruction pipeline, 2) each reprojected input image $\arr{I}^\prime_i$, and 3) a view similarity score $s$ based on the angle between the input and query poses.


\subsection{Implementation Details}

\myparagraph{Training Data} We train our reconstruction network on shapes from ShapeNet~\cite{chang2015shapenet} which contains around 51,300 shapes. We exclude large models for efficient data loading resulting in around 30,000 models. We generate UV maps using Blender's smart UV projection~\cite{blender2019} to facilitate texturing. We normalize all models to unit diameter. When rendering, we sample a random image from MS-COCO~\cite{lin2014microsoft} for each component of the model. We render with the Beckmann model~\cite{beckmann1987scattering} with randomized parameters and also render uniformly colored objects with a probability of 0.5.

\myparagraph{Network Input} We generate our training data at a resolution of $640\times 480$. However, the input to our network is a fixed size $128\times 128$. To keep our inputs consistent and our network scale-invariant, we `zoom' into the object such that all images appear to be from the same distance. This is done by computing a bounding box size $(w_b,h_b)=(\frac{d^\prime w^\prime}{f_u d w},\frac{d^\prime h^\prime}{f_v d h})$ where $(w,h)$ is the current image width and height, $d$ is the distance to the centroid $c_o$ (See Fig.~\ref{fig:camera_frustum}), $(w^\prime,h^\prime)$ is the desired output size, and $d^\prime$ is the desired `zoom' distance and cropping around object centroid projected to image coordinates $(c_u,c_v)$. This defines the viewport parameter $\mat{c}=(c_u-w_b/2,c_v-h_b/2,c_u+w_b/2,c_v+h_b/2)$. The cropped image is scaled to $128\times 128$.

\myparagraph{Training} In each iteration of training, we sample a 3D model and then sample 16 random reference poses and 16 random target poses. Each pose is sampled by uniformly sampling a unit quaternion and translation such that the object stays within frame. We train our network using the Adam optimizer~\cite{kingma2014adam} with a fixed learning rate of 0.001 for 1.5M iterations. Each batch consists of 20 objects with 16 input views and 16 target views. We use an $L_1$ reconstruction loss for depth and binary cross-entropy for the mask. We apply the losses to both the input and output views. We randomly orient our canonical coordinate frame in each iteration by uniformly sampling a random unit quaternion. This prevents our network from overfitting to the implementation of our latent voxel transformations. We also add motion blur, color jitter, and pixel noise to the color inputs and add noise to the input masks using the same procedure as~\cite{mahler2017dex}.

\section{Object Pose Estimation}
\label{sec:pose}

Given an image $\arr{I}$, and a depth map $\arr{D}$, a pose estimation system provides a rotation $\mat{R}$ and a translation $\mat{t}$ which together define an object-to-camera coordinate transformation $\mat{E}=[\mat{R}|\mat{t}]$ referred to as the \emph{object pose}. In this section, we describe how we use our reconstruction pipeline described in Sec.~\ref{sec:recon} to directly optimize for the pose. We first find a \emph{coarse} pose using only forward inference and then \emph{refine} it using gradient optimization.

\myparagraph{Formulation} Pose is defined by a rotation $\mat{R}$ and a translation $\mat{t}$. Our formulation also includes the viewport parameter $\mat{c}$ defined in Sec.~\ref{sec:camera_model}. Defining a viewport allows us to efficiently pass the input to the reconstruction network while also providing scale invariance. We encode the rotation as a quaternion $\mat{q}$ and translation as $\mat{t}$. We assume we are given an RGB image $\arr{I}$, an object segmentation mask $\arr{M}$, and depth $\arr{D}$ comprising the input $\mat{x} = \{\arr{I},\arr{M},\arr{D}\}$.

\subsection{Losses}
In order to estimate pose, we must provide a criterion which quantifies the quality of the pose. We use four loss functions. One is a standard $L_1$ depth reconstruction loss $\mathcal{L}_\text{depth}(\arr{D^*}, \arr{D}) = \left\lVert \arr{D^*} - \arr{D} \right\rVert_1$
which disambiguates the object scale and measures how well the predicted depth $\arr{D}$ matches the input depth $\arr{D^*}$. We also use a pixel-wise binary cross entropy loss $\mathcal{L}_\text{mask}$ on the predicted mask as well as intersection over union (IoU) loss $\mathcal{L}_\text{iou}(\arr{M^*},\arr{M}) = \log{U} - \log{I}$ where $U$ is the sum of the pixels in the union and $I$ is the sum of the pixels in the intersection of the masks $\arr{M^*}$ and $\arr{M}$.  Finally, we introduce a novel latent loss which leverages our reconstruction network $F$. Given the input $\mat{x} = \{\arr{I},\arr{M},\arr{D}\}$, a latent object $\mat{\Psi}$, and a pose $\mat{\theta}$, the latent loss is defined as
    $\mathcal{L}_\text{latent}(\mat{x},\mat\theta;\mat\Psi) = \left\lVert H_{\mat\theta}(G_{\mat\theta}(\mat{x})) - H_{\mat\theta}(\mat\Psi) \right\rVert_1$,
where $H_{\mat\theta}$ is the rendering network up to the projection layer and $G_{\mat\theta}$ is the modeling network as described in Sec.~\ref{sec:recon}. This loss differs from auto-encoder based approaches such as \cite{sundermeyer2018implicit,Deng2019} in that 1) our network is not trained on the object, and 2) the loss is computed directly given the image and camera pose. Our pose estimation problem is given by:
\begin{align}
    \argmin_{\mat\theta}{  
    \mathcal{L}_\text{depth}
    + \lambda\mathcal{L}_\text{latent} 
    + \gamma\mathcal{L}_\text{mask} 
    + \eta\mathcal{L}_\text{iou} } \label{eq:tot-loss},
\end{align}
where $\lambda,\gamma,\eta$ are the weights of the lossees. The parameters of the losses are omitted for clarity.

\myparagraph{Parameterization} We parameterize the rotation in the log quaternion form $\mat{\omega}=(0, \omega_1, \omega_2, \omega_3)$ which ensures that all updates to the parameters result in a valid unit quaternion:
\begin{align}
    \mat{q} &= \exp{(\mat{\omega})} = \begin{pmatrix}
        \cos\lVert\mat{\omega}\rVert \\
        \frac{\mat{\omega}}{\lVert \mat{\omega}\rVert}
        \sin{ \lVert \mat{\omega}\rVert }
    \end{pmatrix}.
\end{align}

\myparagraph{Coarse Initialization} Although we have a differentiable renderer, the space of poses is non-convex which can lead to bad local minima when using gradient based optimization. We therefore bootstrap the pose by computing a coarse estimate. This also has the benefit of speeding up inference since it only requires forward evaluation. 

We begin by estimating the translation of the object $\mat{t}=(x, y, z)$ as the centroid of the bounding cube defined by the mask bounding box $\mat{c}$ and corresponding depth values.  We initialize $k$ poses using the estimated translation. To get good coverage of possible orientations, we evenly sample azimuth and elevation angles using a Fibonacci lattice~\cite{gonzalez2010measurement} then uniformly sample a random yaw angle. We use the cross entropy method~\cite{de2005tutorial} to optimize the translation and log quaternion parameters, and use a Gaussian Mixture Model as the probability distribution.


\myparagraph{Pose Optimization} Our entire pipeline is differentiable end-to-end. We can therefore optimize Eq.~\eqref{eq:tot-loss} using gradient optimization. Given a latent object $\mat{\Psi}$ and a coarse pose estimate $\mat\theta$, we compute the loss and propagate the gradients back to the camera pose. This step only requires the rendering network and does not use the modeling network. The image-based rendering network is also not used in this step. We jointly optimize the rotation $\mat{q}$, translation $\mat{t}$, and viewport $\mat{c}$ using Adam~\cite{kingma2014adam}.

\begin{figure}[t]
\centering
\includegraphics*[width = \linewidth]{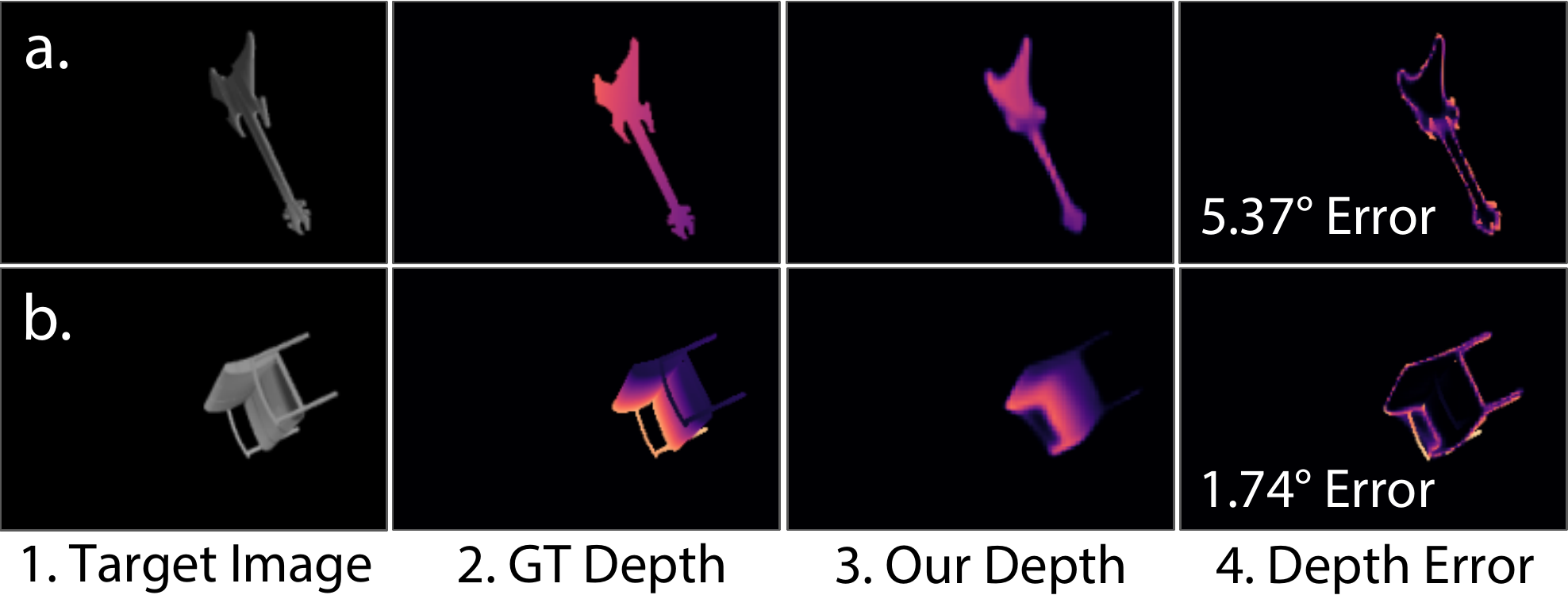}
\vspace{-6mm}
\caption{Two examples from the ModelNet experiment. (1) target image, (2) ground truth depth, (3) optimized predicted depth, and (4) $L_1$ error between the ground truth and our prediction. (a) illustrates how a pose with low depth error can still result in a relatively high angular error.}
\label{fig:modelnet_figure}
\end{figure}

\begin{figure}[t]
\centering
\includegraphics*[width = \linewidth]{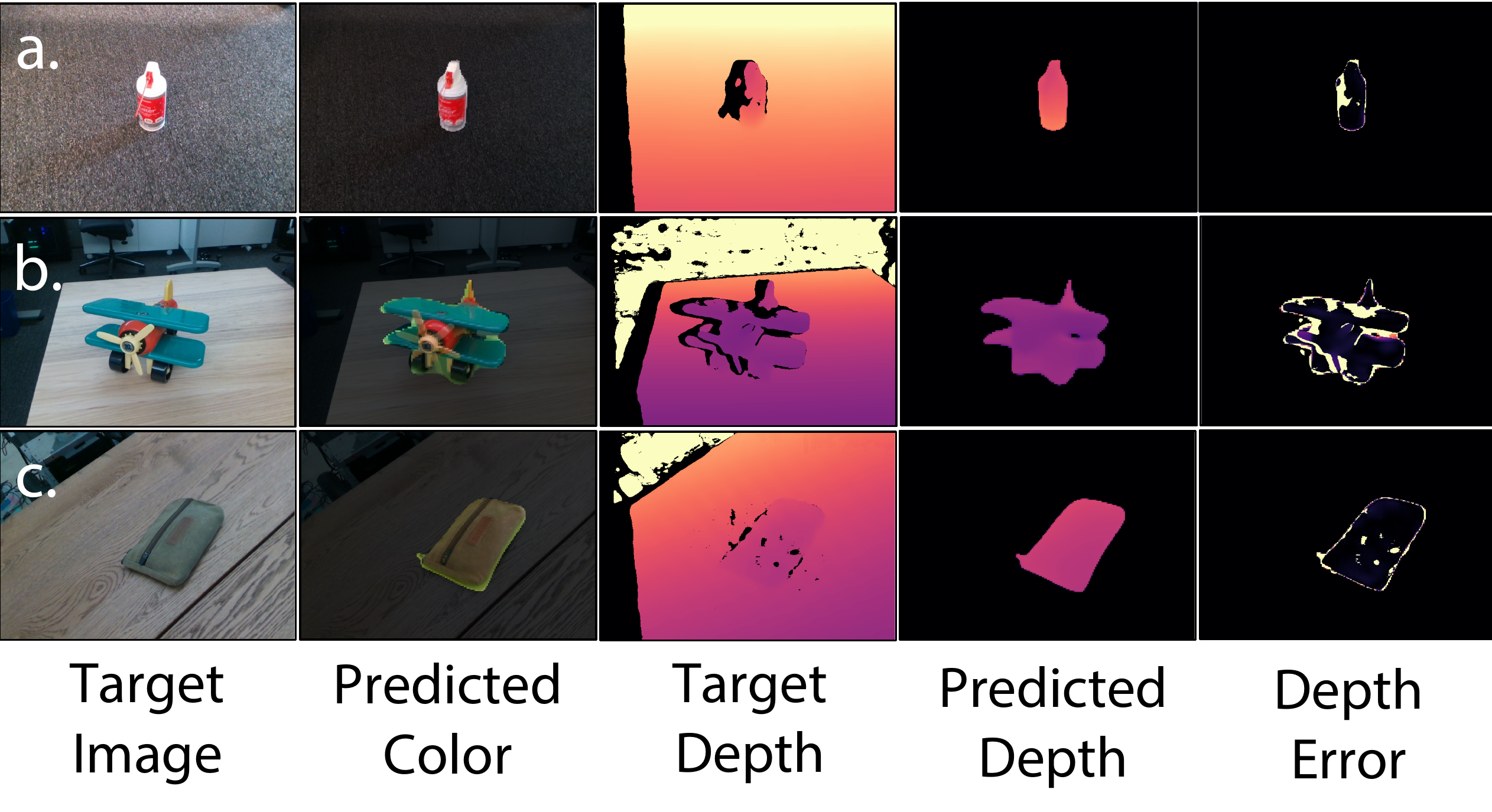}
\vspace{-8mm}
\caption{Qualitative results on the MOPED dataset.}
\label{fig:moped_results}
\vspace{-2mm}
\end{figure}

\begin{table*}[]
\caption{Evaluation on LINEMOD. We report the ADD recall metric~\cite{hinterstoisser2012model}, comparing against DeepIM and pix2pose. Symmetric objects are indicated by $^*$ and the pose is considered correct if it is flipped along the $z$ axis.}
\label{tbl:linemod-pose}
\centering
\scalebox{0.80}{
\setlength\tabcolsep{5pt} 
\begin{tabular}{l|l|cccccccccccccc}
\toprule
Method   & Input & ape           & benchvice     & camera        & can           & cat           & driller       & duck          & eggbox$^*$  & glue$^*$    & holepunch   & iron          & lamp          & phone         & mean          \\ \hline
\textbf{Ours}     & RGB-D & \textbf{88.0} & 92.4          & 74.4          & 88.8          & \textbf{94.5} & 91.7          & 68.1          & 96.3          & 94.9          & \textbf{82.1}          & 74.6          & 94.7          & \textbf{91.5} & 87.1          \\
DeepIM~\cite{Li2018}   & RGB   & 77.0          & \textbf{97.5} & \textbf{93.5} & \textbf{96.5} & 82.1          & \textbf{95.0} & \textbf{77.7} & \textbf{97.1} & \textbf{99.4} & 52.8          & \textbf{98.3} & \textbf{97.5} & 87.7          & \textbf{88.6} \\
pix2pose~\cite{park2019pix2pose} & RGB   & 58.1          & 91.0          & 60.9          & 84.4          & 65.0          & 76.3          & 43.8          & 96.8          & 79.4          & 74.8 & 83.4          & 82.0          & 45.0          & 72.4          \\
\bottomrule
\end{tabular}}
\vspace{-3mm}
\end{table*}

\section{Experiments}
We evaluate our method on LINEMOD~\cite{hinterstoisser2011multimodal}, ModelNet~\cite{wu20153d} and our new dataset MOPED. We aim to evaluate pose estimation accuracy on unseen objects.

\subsection{Evaluation Metrics}
We use four main evaluation metrics. 1) $(k^\circ, k\text{cm})$: a pose is considered correct if it is within $k^\circ$ and $k\text{cm}$ of the ground truth target pose where the angular metric is the angle between the orientations. 2) ADD~\cite{hinterstoisser2012model}: the average distance between points after being transformed by the ground truth and predicted poses. 3) ADD-S: A modification to ADD that computes the average distance to the closest point rather than the ground truth point to account for symmetric objects. 4) Proj.2D: the pixel distance between the projected points of the ground truth and predicted pose.

\subsection{Experiments on the LINEMOD Dataset}
We evaluate our method on the LINEMOD dataset. We compare our results with DeepIM~\cite{Li2018} and pix2pose~\cite{park2019pix2pose}. Both of these methods are trained on the LINEMOD dataset. DeepIM uses the 3D models in an online fashion to refine the pose. We \emph{do not train} on the dataset. Instead, our network is given 16 reference views of each object during inference time. We use the provided segmentation masks during inference. We follow the train/test split of \cite{hodan2018bop} and sample our reference views from the training set. We follow the same evaluation methodology of \cite{hinterstoisser2012model}, reporting the percentage of poses with ADD metric less than 10\% of the object diameter. Table.~\ref{tbl:linemod-pose} shows the results. Our experiments show that our method performs on-par with state of the art supervised methods despite having never seen the objects.

\subsection{Experiments on the ModelNet Dataset}
We conduct experiments on ModelNet~\cite{wu20153d} to evaluate the generalization of our method toward unseen object categories. To this end, we train our network on all the meshes in ShapeNetCore~\cite{chang2015shapenet} excluding the categories we are going to evaluate on. We closely follow the evaluation protocol of \cite{Li2018} here. The model is evaluated on 7 unseen categories: {\it bathtub, bookshelf, guitar, range hood, sofa, wardrobe}, and {\it TV stand}. For each category, 50 pairs of initial and target object pose are sampled. We compared with \cite{Li2018} and  \cite{sundermeyer2019multi} where all the methods initialized with the initial pose and evaluated on how successful they are on estimating the target pose. We report three metrics: $(5^\circ, 5cm)$, \emph{ADD} within 10\% of the object diameter, and \emph{Proj.2D} within 5 pixels.

Table~\ref{tbl:modelnet-pose} shows the quantitative results on the ModelNet dataset. On average, our method achieves state-of-the-art results on all the metrics thanks to our ability to perform continuous optimization on pose. However, for the  $(5^\circ, 5cm)$ metric, there are object categories that our method performs worse despite performing well on all other metrics. One reason is the image and spatial resolution. The input and output images to our network have resolution $128\times128$. The resolution of our voxel representation is $16 \times 16 \times 16$. The limited resolution can hinder the performance for small objects or objects that are distant from the camera. Small changes in the depth of each pixel may disproportionately affect the rotation of the object compared to our losses. Fig.~\ref{fig:modelnet_figure} shows examples from the ModelNet experiment illustrating this limitation.

\begin{table}

\caption{ModelNet pose refinement experiments compared to DeepIM (DI)~\cite{Li2018} and Multi-Path Learning (MP)~\cite{sundermeyer2019multi}.}
\label{tbl:modelnet-pose}
\scalebox{0.75}{
\setlength\tabcolsep{5pt} 
\begin{tabular}{l|ccc|ccc|ccc}
\toprule
            & \multicolumn{3}{c|}{(5°, 5cm)}                                              & \multicolumn{3}{c|}{ADD (0.1d)}                                              & \multicolumn{3}{c}{Proj2D (5px)}                                          \\ \hline
            & \multicolumn{1}{c}{DI} & \multicolumn{1}{c}{MP} & \multicolumn{1}{c|}{Ours} & \multicolumn{1}{c}{DI} & \multicolumn{1}{c}{MP} & \multicolumn{1}{c|}{Ours}   & \multicolumn{1}{c}{DI} & \multicolumn{1}{c}{MP} & \multicolumn{1}{c}{Ours} \\ \hline
bathtub     & 71.6                   & \textbf{85.5}          & 85.0                      & 88.6                   & 91.5                   & \textbf{92.7}             & 73.4                   & 80.6                   & \textbf{94.9}            \\
bookshelf   & 39.2                   & \textbf{81.9}          & 80.2                      & 76.4                   & 85.1                   & \textbf{91.5}             & 51.3                   & 76.3                   & \textbf{91.8}            \\
guitar      & 50.4                   & 69.2                   & \textbf{73.5}             & 69.6                   & 80.5                   & \textbf{83.9}                   & 77.1                   & 80.1                   & \textbf{96.9}            \\
range\_hood & 69.8                   & \textbf{91.0}          & 82.9                      & 89.6                   & 95.0                   & \textbf{97.9}             & 70.6                   & 83.9                   & \textbf{91.7}            \\
sofa        & 82.7                   & \textbf{91.3}          & 89.9                      & 89.5                   & 95.8                   & \textbf{99.7}                    & 94.2                   & 86.5                   & \textbf{97.6}            \\
tv\_stand   & 73.6                   & 85.9                   & \textbf{88.6}             & 92.1                   & 90.9                   & \textbf{97.4}             & 76.6                   & 82.5                   & \textbf{96.0}            \\
wardrobe    & 62.7                   & 88.7                   & \textbf{91.7}             & 79.4                   & 92.1                   & \textbf{97.0}    &          70.0                   & 81.1                   & \textbf{94.2}            \\ \hline
Mean        & 64.3                   & 84.8                   & \textbf{85.5}             & 83.6                   & 90.1                   & \textbf{94.3}             & 73.3                   & 81.6                   & \textbf{94.7}            \\ 
\bottomrule
\end{tabular}}
\vspace{-2mm}
\end{table}

\begin{table}
\caption{AUC metrics on MOPED by reference view count.}
\label{tbl:num_views}
\centering
\scalebox{0.75}{\begin{tabular}{c|cccccc}
\toprule 
\# Views & 1     & 2     & 4     & 8     & 16    & 32    \\
\hline
ADD             & 23.9	& 34.2 &	49.8 &	{\bf 63.1} &	62.7 &	61.1 \\
ADD-S           & 78.7 &	82.8 &	89.2 &	{\bf 91.1} &	90.8 &	90.8 \\
Proj.2D          & 9.5 &	15.9 &	30.4 &	44.7 &	{\bf 46.8} &	42.6 \\
\bottomrule
\end{tabular}}
\end{table}

\begin{table}
\centering
\caption{AUC metrics for different view fusion strategies}
\label{tbl:view_fusion}
\scalebox{0.75}{
\begin{tabular}{c|ccc}
\toprule
         & ADD   & ADD-S & Proj.2D \\
\hline
Avg Pool & 57.5 & 90.1 & 40.1  \\
ConvGRU  & {\bf 63.1} & {\bf 91.1} & {\bf 44.7} \\
\bottomrule
\end{tabular}}
\end{table}

\begin{table*}[ht]
\centering
\caption{Quantitative Results on MOPED Dataset. We report the Area Under Curve (AUC) for each metric. $^{\dagger}$ shows the average with the symmetric objects rinse\_aid and cheezit excluded. Ours (D) is our method with the depth loss only and Ours (D+L) is our method with both the depth and latent losses. In some cases, the latent loss helps resolve pose ambiguities not possible with only depth.}
\label{tab:moped-results}
\scalebox{1.0}{
\begin{tabular}{l|lll|lllrrr}
\toprule
               & \multicolumn{3}{c|}{PoseRBPF}                                                                 & \multicolumn{3}{c|}{Ours (D)}                                                                 & \multicolumn{3}{c}{Ours (D+L)}                                                                                                    \\ \hline
Input          & \multicolumn{3}{c|}{Textured 3D Mesh}                                                         & \multicolumn{6}{c}{\textbf{Image + Camera Pose}}                                                                                                                                                                                 \\ \hline
Training       & \multicolumn{3}{c|}{Yes}                                                                      & \multicolumn{6}{c}{\textbf{No}}                                                                                                                                                                                                  \\ \hline
\# Networks    & \multicolumn{3}{c|}{Per-Object}                                                               & \multicolumn{6}{c}{\textbf{Single Universal}}                                                                                                                                                                                    \\ \hline\hline
Object         & \cellcolor[HTML]{B7B7B7}ADD           & \cellcolor[HTML]{D9D9D9}ADD-S         & Proj.2D       & \cellcolor[HTML]{B7B7B7}ADD           & \cellcolor[HTML]{D9D9D9}ADD-S         & Proj.2D       & \multicolumn{1}{l}{\cellcolor[HTML]{B7B7B7}ADD} & \multicolumn{1}{l}{\cellcolor[HTML]{D9D9D9}ADD-S} & \multicolumn{1}{l}{Proj.2D} \\ \hline
black\_drill   & \cellcolor[HTML]{B7B7B7}75.3          & \cellcolor[HTML]{D9D9D9}91.7          & 54.2          & \cellcolor[HTML]{B7B7B7}\textbf{90.3} & \cellcolor[HTML]{D9D9D9}\textbf{95.4} & \textbf{78.1} & \cellcolor[HTML]{B7B7B7}89.4                    & \cellcolor[HTML]{D9D9D9}95.1                      & 76.2                        \\ \hline
duplo\_dude    & \cellcolor[HTML]{B7B7B7}84.1          & \cellcolor[HTML]{D9D9D9}93.5          & 61.5          & \cellcolor[HTML]{B7B7B7}\textbf{89.4} & \cellcolor[HTML]{D9D9D9}\textbf{94.9} & \textbf{77.4} & \cellcolor[HTML]{B7B7B7}88.9                    & \cellcolor[HTML]{D9D9D9}94.7                      & 76.2                        \\ \hline
duster         & \cellcolor[HTML]{B7B7B7}\textbf{72.2} & \cellcolor[HTML]{D9D9D9}\textbf{90.6} & \textbf{51.5} & \cellcolor[HTML]{B7B7B7}40.0          & \cellcolor[HTML]{D9D9D9}88.3          & 17.8          & \cellcolor[HTML]{B7B7B7}47.2                    & \cellcolor[HTML]{D9D9D9}87.9                      & 15.5                        \\ \hline
graphics\_card & \cellcolor[HTML]{B7B7B7}71.8          & \cellcolor[HTML]{D9D9D9}\textbf{81.3} & 51.2          & \cellcolor[HTML]{B7B7B7}\textbf{73.1} & \cellcolor[HTML]{D9D9D9}79.3          & 62.2          & \cellcolor[HTML]{B7B7B7}73.0                    & \cellcolor[HTML]{D9D9D9}79.2                      & \textbf{62.5}               \\ \hline
orange\_drill  & \cellcolor[HTML]{B7B7B7}70.3          & \cellcolor[HTML]{D9D9D9}87.5          & 43.9          & \cellcolor[HTML]{B7B7B7}\textbf{78.4} & \cellcolor[HTML]{D9D9D9}\textbf{93.7} & \textbf{62.3} & \cellcolor[HTML]{B7B7B7}78.3                    & \cellcolor[HTML]{D9D9D9}93.5                      & 61.7                        \\ \hline
pouch          & \cellcolor[HTML]{B7B7B7}26.5          & \cellcolor[HTML]{D9D9D9}80.7          & 14.1          & \cellcolor[HTML]{B7B7B7}\textbf{61.9} & \cellcolor[HTML]{D9D9D9}\textbf{91.8} & \textbf{44.2} & \cellcolor[HTML]{B7B7B7}54.8                    & \cellcolor[HTML]{D9D9D9}91.7                      & 31.4                        \\ \hline
remote         & \cellcolor[HTML]{B7B7B7}49.6          & \cellcolor[HTML]{D9D9D9}82.1          & 25.1          & \cellcolor[HTML]{B7B7B7}58.2          & \cellcolor[HTML]{D9D9D9}\textbf{92.0} & 19.0          & \cellcolor[HTML]{B7B7B7}\textbf{60.0}           & \cellcolor[HTML]{D9D9D9}91.9                      & \textbf{26.3}               \\ \hline
toy\_plane     & \cellcolor[HTML]{B7B7B7}48.6          & \cellcolor[HTML]{D9D9D9}88.6          & 29.5          & \cellcolor[HTML]{B7B7B7}55.5          & \cellcolor[HTML]{D9D9D9}89.7          & 46.5          & \cellcolor[HTML]{B7B7B7}\textbf{82.9}           & \cellcolor[HTML]{D9D9D9}\textbf{92.2}             & \textbf{70.5}               \\ \hline
vim\_mug       & \cellcolor[HTML]{B7B7B7}\textbf{59.0} & \cellcolor[HTML]{D9D9D9}\textbf{92.9} & \textbf{39.4} & \cellcolor[HTML]{B7B7B7}51.5          & \cellcolor[HTML]{D9D9D9}91.8          & 22.7          & \cellcolor[HTML]{B7B7B7}40.0                    & \cellcolor[HTML]{D9D9D9}91.7                      & 8.0                         \\ \hline
rinse\_aid     & \cellcolor[HTML]{B7B7B7}87.2          & \cellcolor[HTML]{D9D9D9}\textbf{94.7} & \textbf{68.7} & \cellcolor[HTML]{B7B7B7}\textbf{71.3} & \cellcolor[HTML]{D9D9D9}93.2          & 41.7          & \cellcolor[HTML]{B7B7B7}67.4                    & \cellcolor[HTML]{D9D9D9}93.3                      & 34.7                        \\ \hline
cheezit        & \cellcolor[HTML]{B7B7B7}\textbf{64.3} & \cellcolor[HTML]{D9D9D9}\textbf{93.8} & \textbf{55.6} & \cellcolor[HTML]{B7B7B7}24.8          & \cellcolor[HTML]{D9D9D9}92.4          & 20.5          & \cellcolor[HTML]{B7B7B7}24.6                    & \cellcolor[HTML]{D9D9D9}92.2                      & 20.2                        \\ \hline\hline
Mean           & \cellcolor[HTML]{B7B7B7}\textbf{64.4} & \cellcolor[HTML]{D9D9D9}88.9          & \textbf{45.0} & \cellcolor[HTML]{B7B7B7}63.1          & \cellcolor[HTML]{D9D9D9}91.1          & 44.7          & \cellcolor[HTML]{B7B7B7}64.2                    & \cellcolor[HTML]{D9D9D9}\textbf{91.2}             & 43.9                        \\ \hline
Mean w/o sym   & \cellcolor[HTML]{B7B7B7}61.9          & \cellcolor[HTML]{D9D9D9}87.7          & 41.2          & \cellcolor[HTML]{B7B7B7}66.5          & \cellcolor[HTML]{D9D9D9}90.8          & 47.8          & \cellcolor[HTML]{B7B7B7}\textbf{68.3}           & \cellcolor[HTML]{D9D9D9}\textbf{90.9}             & \textbf{47.6}               \\ \bottomrule
\end{tabular}}

\vspace{-2mm}
\end{table*}

\begin{figure}[t]
\centering
\includegraphics*[width = \linewidth]{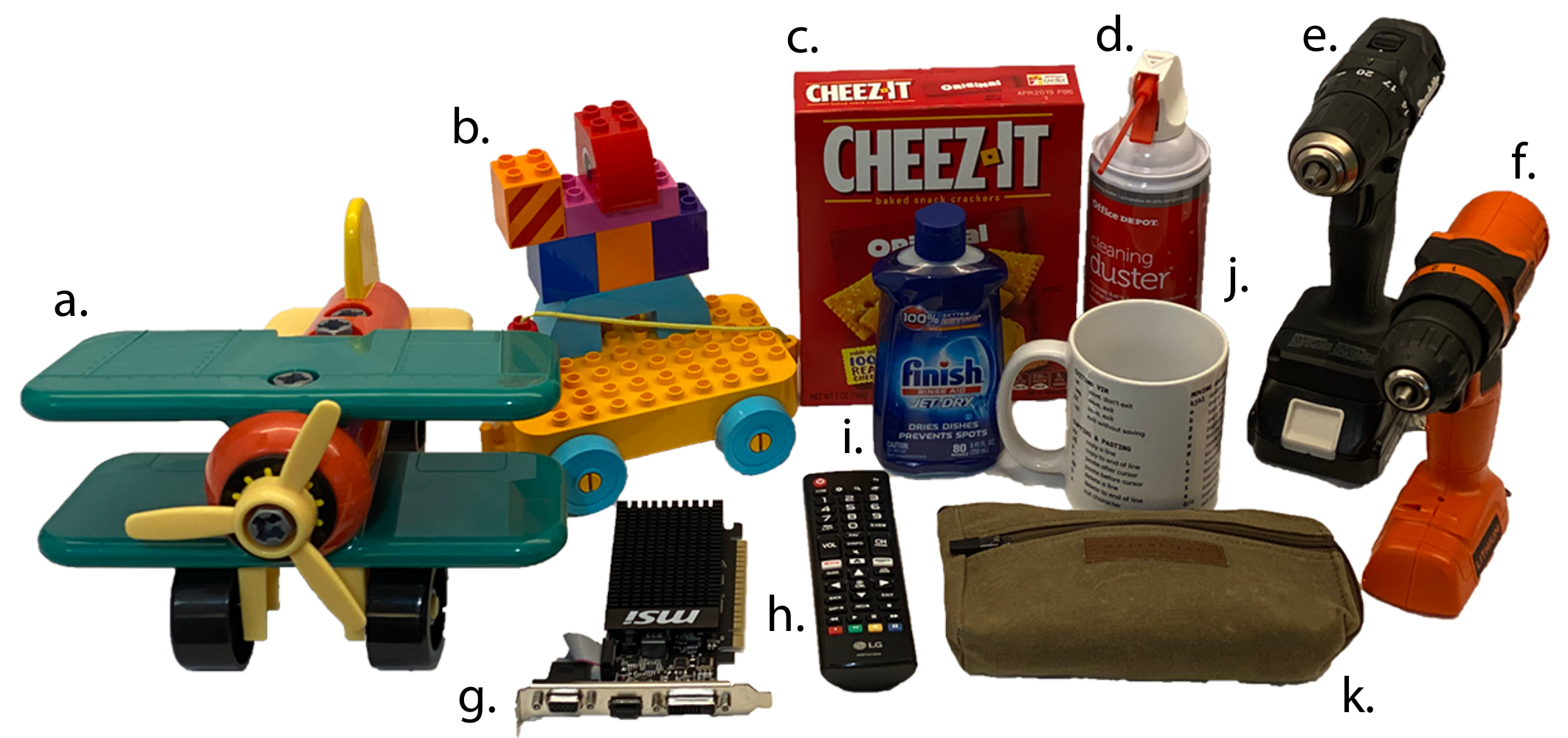}
\vspace{-7mm}
\caption{Objects in MOPED--a new dataset for model-free pose estimation. The objects shown are: (a) toy\_plane, (b) duplo\_dude, (c) cheezit, (d) duster, (e) black\_drill, (f) orange\_drill, (g) graphics\_card, (h) remote, (i) rinse\_aid, (j) vim\_mug, and (k) pouch.}
\label{fig:moped_overview}
\vspace{-5mm}
\end{figure}

\subsection{Experiments on the MOPED Dataset}
We introduce the Model-free Object Pose Estimation Dataset (MOPED). MOPED consists of 11 objects, shown in Fig.~\ref{fig:moped_overview}. For each object, we take multiple RGB-D videos cover all views the object. We first use KinectFusion~\cite{newcombe2011kinectfusion} to register frames from a single capture and then use a combination of manual annotation and automatic registration~\cite{zhou2016fast,rusinkiewicz2001efficient,Zhou2018} to align separate captures. We generate object segmentation maps using~\cite{Xie2019}. For each object, we select reference frames with farthest point sampling to ensure good coverage of the object. For test sequences, we capture each object in 5 different environments. We sample every other frame for evaluation videos. This results in approximately 300 test images per object. We evaluate our method and baselines using three metrics for which we provide the Area Under Curve (AUC): 1) \emph{ADD} with threshold between $0-10cm$, 2) \emph{ADD-S} with threshold between $0-10cm$, and 3) \emph{Proj.2D} with threshold between $0-40px$. We compute all metrics for all sampled frames.

We compare our method with PoseRBPF~\cite{Deng2019}, a state-of-the-art model-based pose estimation method. Since PoseRBPF requires textured 3D models, we reconstruct a mesh for each object by aggregating point clouds from reference captures and building a TSDF volume. The point clouds are integrated into the volume using KinectFusion~\cite{newcombe2011kinectfusion}. The meshes have artifacts such as washed out high frequency details and shadow vertices due to slight misalignment (see supplementary materials). Table~\ref{tab:moped-results} shows quantitative comparisons on the MOPED dataset. Note that our method is not trained on the test objects while PoseRBPF has a separate encoder for each object. Our method achieves superior performance on both ADD and ADD-S. We evaluate different version of our method with different combinations of loss functions. Compared to our combined loss, optimizing only $\mathcal{L}_\text{depth}$ performs better for geometrically asymmetric objects  but worse on textured  objects such as the cheezit box. Optimizing both losses achieves better results on textured objects. Fig.~\ref{fig:moped_results} shows estimated poses for different test images. Please see supplementary materials for qualitative examples.



\subsection{Ablation Studies}
\label{sec:ablation}
In this section, we analyze the effect of different design choices and how they affect the robustness of our method.

\vspace{1mm}
\myparagraph{Number of reference views} We first evaluate the sensitivity of our method to the number of input reference views. Novel view synthesis is easier with  more reference views because there is a higher chance that a query view will be close to a reference view. Table~\ref{tbl:num_views} shows that the accuracy increases with the number of reference views. In addition, having more than 8 reference views only yields marginal performance gains shows that our method does not require many views to achieve good pose estimation results.

\myparagraph{View Fusion} We compare multiple strategies for aggregating the latent representations from each reference view. The naive way is to use a simple pooling function such as average/max  pooling. Alternatively, we can integrate the volumes using an RNN such as a ConvGRU so that the network can reason across views. Table~\ref{tbl:view_fusion}  shows the quantitative evaluation of these two variations. Although the average performance of the objects are very similar, the ConvGRU variation performs better than the average pooling variation. This indicates the importance of spatial relationship in the voxel representation for pose estimation.

\section{Conclusion}

We have presented a novel framework for building 3D object representations at inference time using a small number of reference images, as well as an accompanying neural renderer to render the 3D representation from arbitrary 6D viewpoints. Our networks are trained on thousands of shapes with random textures rendered under various lighting conditions allowing it to robustly generalize to unseen objects \emph{without additional training}.  

We leverage our reconstruction and rendering pipeline for zero-shot pose estimation. We perform pose estimation given just a small number of reference views and without needing to train any network. This greatly simplifies the process for performing pose estimation on novel objects as a detailed 3D model is not required. In addition, we have a single universal network which works for all objects including unseen ones. For future work, we plan to investigate unseen object pose estimation in cluttered scenes with occlusions. We also plan to speed up the pose estimation process by applying network optimization techniques.


\section*{Acknowledgements}
We thank Xinke Deng for helpful discussions.

{\small
\bibliographystyle{ieee_fullname}
\bibliography{references}
}

\end{document}